\documentclass[12pt,a4paper]{article}

\usepackage[margin=1in]{geometry}
\usepackage[english]{babel}

\usepackage{amsmath}   
\usepackage{amssymb}   
\usepackage{amsthm}    
\usepackage{bm}        
\usepackage{mathrsfs}  
\usepackage{physics}   
\usepackage{nicefrac}  
\usepackage{gensymb}   

\usepackage{graphicx}
\usepackage{epstopdf}  
\usepackage{wrapfig}
\usepackage{longtable}
\usepackage{booktabs}
\usepackage{array}
\usepackage{multirow}
\usepackage{adjustbox}
\usepackage[table]{xcolor} 

\usepackage[numbers]{natbib} 

\usepackage{enumitem}

\usepackage{csquotes}   
\usepackage{comment}    
\usepackage{listings}   
\usepackage{hyperref}   

\usepackage{tikz}
\usetikzlibrary{matrix,fit,positioning,decorations.pathreplacing,arrows,
                calc,decorations.pathmorphing,patterns,shapes}

\title{\LARGE PCA-RAG: Principal Component Analysis for Efficient Retrieval-Augmented Generation}

\author{
  Arman Khaledian\thanks{Zanista AI \quad \texttt{arman.khaledian@zanista.ai}}
  \and
  Amirreza Ghadiridehkordi\thanks{Zanista AI \quad \texttt{amir.dehkordi@zanista.ai}}
  \and
  Nariman Khaledian\thanks{Zanista AI \quad \texttt{nariman.khaledian@zanista.ai}}
}

\date{\today}

\begin{document}
\maketitle





\begin{abstract}
Retrieval-Augmented Generation (RAG) has emerged as a powerful paradigm for grounding large language models in external knowledge sources, improving the precision of agents responses. However, high-dimensional language model embeddings—often in the range of hundreds to thousands of dimensions—can present scalability challenges in terms of storage and latency, especially when processing massive financial text corpora. This paper investigates the use of Principal Component Analysis (PCA) to reduce embedding dimensionality, thereby mitigating computational bottlenecks without incurring large accuracy losses. We experiment with a real-world dataset and compare different similarity and distance metrics under both full-dimensional and PCA-compressed embeddings. Our results show that reducing vectors from 3{,}072 to 110 dimensions provides a sizeable (up to \(60\times\)) speedup in retrieval operations and a \(\sim 28.6\times\) reduction in index size, with only moderate declines in correlation metrics relative to human-annotated similarity scores. These findings demonstrate that PCA-based compression offers a viable balance between retrieval fidelity and resource efficiency, essential for real-time systems such as Zanista AI’s \textit{Newswitch} platform. Ultimately, our study underscores the practicality of leveraging classical dimensionality reduction techniques to scale RAG architectures for knowledge-intensive applications in finance and trading, where speed, memory efficiency, and accuracy must jointly be optimized.
\end{abstract}

\section{Introduction}
Recent advances in large language models (LLMs) and retrieval-based systems have paved the way for new applications in finance and trading, where swift access to accurate, domain-specific information is critical \cite{brown2020, karpukhin2020}. At Zanista AI, we have developed \textit{Newswitch}, a platform designed to serve real-time news retrieval and sentiment analysis in the financial sector. \textit{Newswitch} leverages Retrieval-Augmented Generation (RAG) to ground AI Investment Agent responses on up-to-date market data, enabling more precise and context-aware answers to financial queries \cite{lewis2020, izacard2020}.

Yet, a well-known challenge in RAG pipelines is the high-dimensional nature of the embedding vectors used for retrieval. Models like BERT and other Transformer-based encoders often produce 768- to 3,000-dimensional vectors \cite{devlin2019, openai2023}, creating large storage requirements and incurring significant computational costs---both of which are amplified when dealing with massive financial text corpora such as news articles and regulatory filings \cite{iaroshev2024}.

To address this challenge, researchers have turned to dimensionality reduction techniques, with Principal Component Analysis (PCA) emerging as a simple yet powerful tool \cite{ma2021, yang2021}. By projecting high-dimensional embeddings onto a lower-dimensional subspace, PCA significantly reduces index size and query latency, often with minimal impact on retrieval accuracy \cite{liu2022}. In finance-focused applications such as \textit{Newswitch}, where speed and scale are paramount, PCA thus presents an attractive trade-off between retaining semantic fidelity in embeddings and managing practical constraints like memory and throughput.

In this paper, we examine how PCA-based compression can improve the efficiency of RAG pipelines for retrieval and question answering. We draw on prior work indicating that most variance in dense embeddings can be preserved with a modest fraction of dimensions \cite{ma2021}. By assessing the trade-off between retrieval accuracy and computational performance---both in terms of latency and memory footprint---we show that PCA maintains robust alignment with human judgments of semantic similarity while dramatically cutting storage requirements. Taken together, these findings underscore how Zanista AI’s \textit{Newswitch} platform is more effectively integrating RAG techniques at scale, ultimately enhancing Investment Agent's responsiveness in real-time financial decision-making scenarios.

\section{Relevant Work}

\subsection{Principal Component Analysis in Retrieval-Augmented Generation (RAG)}
Retrieval-Augmented Generation (RAG) combines information retrieval with text generation to ground large language models on external knowledge \cite{lewis2020, guu2020}. A key challenge in RAG is efficiency – both in retrieving relevant documents and in processing them during generation \cite{zhu2024}. High-dimensional vector representations (e.g.\ 768-dimensional BERT embeddings) provide strong retrieval accuracy but incur large storage and latency costs \cite{karpukhin2020, izacard2020, xiong2021}. Dimensionality reduction techniques like Principal Component Analysis (PCA) have thus emerged as important tools to compress embeddings, speeding up retrieval and reducing memory usage while preserving accuracy \cite{ma2021, yang2021}.

\subsection{PCA for Efficient Neural Retrieval in NLP}
In natural language processing, PCA and related methods are used to identify and remove redundancy in embedding vectors \cite{deerwester1990, arora2017}. Recent studies show that dense retrievers’ embedding dimensions are often larger than necessary. For example, Ma et al.\ \cite{ma2021} found that a 768-dimensional DPR vector has substantial redundancy – about 99\% of the variance and mutual information is captured in the first $\sim$256 dimensions. They proposed an unsupervised compression pipeline using PCA (followed by product quantization) to shrink embedding size, achieving a 48$\times$ index compression with under 3\% loss in top-100 retrieval accuracy, and even 96$\times$ compression with $<4\%$ drop, all with minimal retrieval effectiveness degradation. Notably, their results showed that unsupervised PCA can match or outperform learned reduction methods at moderate dimensions (128--256). In a similar vein, Yang and Seo \cite{yang2021} demonstrated that using only the first 128 dimensions of 512-d BERT embeddings in an open-domain QA retriever yields a linear 4$\times$ index size reduction with negligible accuracy loss ($\sim$1--2\%).

Beyond PCA, learned compressive models have been explored. Autoencoder approaches can reduce dimensionality while learning to preserve more task-specific features \cite{chen2020, liu2022}. Liu et al.\ \cite{liu2022} introduce a Conditional Autoencoder (ConAE) to compress dense retriever embeddings, achieving nearly the same ranking performance at 128 dimensions as the original 768-d model. For instance, on MS MARCO passage retrieval, compressing embeddings from 768$\rightarrow$128 dims via ConAE only marginally lowered MRR@10 (0.3302 to 0.3245), whereas a naive PCA compression to 128 dims had a larger drop (0.2348). This suggests that while plain PCA finds the directions of maximal variance, learned models can better preserve task-specific signal. Nevertheless, PCA remains a simple and strong baseline for dimensionality reduction in retrieval, often serving as the first step in efficient pipeline designs \cite{reimers2019}. By stripping out low-variance components, PCA tends to remove noise or redundant features in embeddings, which can even improve retrieval robustness in some cases \cite{ma2021,yang2021}.

\subsection{Applications in Finance and Trading Systems}
The finance and trading domain presents massive data – from lengthy regulatory filings to real-time market feeds – where RAG can offer significant value \cite{matsubara2023, chen2023}. Recent work applies RAG to financial document question answering and analytics, leveraging retrieval to inject up-to-date domain knowledge into LLMs (Daizy V2). For instance, Iaroshev et al.\ \cite{iaroshev2024} evaluate RAG for answering questions from annual reports, using a pipeline that chunks documents and encodes them into embeddings stored in a vector database. Such systems typically rely on high-dimensional language model embeddings (e.g.\ 768-dim) to capture fine-grained financial jargon and context. Dimensionality reduction can be especially beneficial here: financial corpora are large, so compressing embeddings reduces index size and query latency, crucial for real-time trading applications. Although specific uses of PCA in finance-focused retrieval are not always singled out, the general principle carries over – fewer dimensions often means faster retrieval – enabling quick lookup of relevant news or reports during trading decisions. Moreover, PCA is a familiar tool in quantitative finance (e.g.\ for factor analysis and risk factor extraction), and analogous use in NLP-based financial systems helps distill essential information from text embeddings. Xiao et al.\ \cite{xiao2025} even extend RAG to time-series forecasting, retrieving historical market patterns to assist an LLM in stock trend prediction.

\subsection{Advances in RAG Architectures and Efficiency}
Beyond embedding compression, recent advances in RAG architectures tackle efficiency at multiple levels. On the retrieval side, approximate nearest neighbor search methods (e.g.\ HNSW graphs, product quantization) are commonly used alongside dimensionality reduction to speed up similarity search in large vector collections \cite{karpukhin2020, johnson2019}. Efficient index structures and caching can mitigate the curse of dimensionality when scaling to billions of vectors. On the generation side, researchers have addressed the inference bottleneck that arises from conditioning on many retrieved documents. A standard RAG model prepends $k$ retrieved passages to the query, causing input length (and thus Transformer cost) to grow linearly with $k$ \cite{izacard2020}. Techniques like Fusion-in-Decoder (FiD) integrate multiple documents by encoding them separately and fusing their representations in the decoder, which improved answer quality for multi-document QA \cite{izacard2020}. However, FiD can be resource-intensive, and follow-up work has sought better speed-quality tradeoffs, such as Parallel Context Windows (PCW) \cite{zhu2024}. More recently, Sparse RAG \cite{zhu2024} introduces a selective attention mechanism: the language model decides which retrieved documents to attend to, dropping others from the context to reduce computation. By only loading the most relevant facts into the decoder, Sparse RAG cuts down on memory and latency, nearly doubling generation speed in experiments without hurting answer accuracy. These architectural innovations are complementary to embedding-level tweaks like PCA. From early RAG systems \cite{lewis2020} to optimized, finance-ready architectures, the trajectory of development reflects a growing need for both retrieval power and responsiveness in LLM-driven applications.

\section{Methodology}

\subsection{Overview of PCA and RAG in Retrieval}
Principal Component Analysis (PCA) is a dimensionality reduction technique that transforms high-dimensional data into a lower-dimensional form while preserving as much variance (information) as possible \cite{jolliffe2002}. By projecting data onto a few principal components (orthogonal directions of maximal variance), PCA can denoise representations and remove redundancies \cite{abdi2010}. In information retrieval, such dimensionality reduction has long been used to capture latent semantics \cite{deerwester1990}. The intuition is that a lower-dimensional representation focuses on the most important features of the text, potentially improving similarity detection by eliminating spurious details \cite{aggarwal2001}.

Retrieval-Augmented Generation (RAG) refers to systems that combine a generative model with an external retrieval mechanism. A RAG system employs a retriever to find relevant documents (from a knowledge base or dataset) given a query, and then feeds those documents into a generator (often a large language model) to produce a final output. This approach allows large language models (LLMs) to augment their internal knowledge with up-to-date or domain-specific information, addressing issues like hallucination and stale knowledge. In practice, RAG implementations use a vector database or index of documents encoded as high-dimensional embeddings, and a similarity search to retrieve the top-$k$ relevant pieces of text for a given query embedding \cite{lewis2020}. For example, Lewis et al.\ introduced RAG by coupling a parametric seq2seq model (the generator) with a non-parametric memory of Wikipedia passages indexed by dense vectors, accessed via a neural retriever. This architecture has proven effective for knowledge-intensive tasks by marrying LLMs’ natural language generation ability with precise, recall-oriented information retrieval.

In the context of our work at Zanista AI – which develops Newswitch, a platform quantifying news and news sentiment for finance and trading – RAG enables the system to retrieve relevant news articles or snippets from a news archive to support real-time analysis. However, the embeddings used for retrieval (for instance, those from Transformer models) are often high-dimensional. This can pose challenges: storing and searching through millions of high-dimension vectors is computationally heavy, and the curse of dimensionality can sometimes degrade retrieval accuracy \cite{aggarwal2001}. By applying PCA to RAG’s embedding space, we aim to retain the essential semantic content needed for retrieval, while reducing vector size and noise.

\subsection{Dataset and Preprocessing}
For our evaluation, we curated a dataset from HuggingFace\footnote{\url{https://huggingface.co/datasets/sentence-transformers/stsb}} sentence pairs with human-annotated similarity scores. Each data point consists of two news sentences (Sentence1 and Sentence2) along with a relevance score in $[0,1]$ indicating their semantic similarity or relatedness. A score of 1 denotes that the two sentences are highly related, whereas 0 means they are entirely unrelated. The total dataset consists of approximately 8{,}600 sentence pairs, split evenly and randomly into training and test sets for experimentation.

\paragraph{Text Embeddings:} We used a state-of-the-art language model to convert each sentence into an embedding vector. In particular, we leveraged \texttt{text-embedding-3-large}, a pretrained transformer-based encoder, to generate a numeric vector for every sentence \cite{openai2023}. The resulting embeddings were high-dimensional (the raw vectors had 3{,}072 dimensions in our case, which is consistent with modern LLM embedding sizes). These embeddings capture semantic meaning – sentences about similar topics or events have closer vectors in this high-dimensional space.

\paragraph{Preprocessing:} Before applying PCA, we performed standard preprocessing on the embeddings. Each embedding was standardized by removing the mean and scaling to unit variance (using \texttt{StandardScaler}). This step ensures that all dimensions are on a comparable scale so that PCA does not overweight dimensions simply due to scale differences. It is a common practice to normalize features prior to PCA, especially since different embedding dimensions could have different value distributions \cite{abdi2010}.

\subsection{Implementation Details (PCA Integration)}
We implemented PCA compression using the \texttt{scikit-learn} library. First, we fit a PCA model on the training set of normalized sentence embeddings. The PCA algorithm computes the principal components of the embedding space. We examined the cumulative explained variance ratio to decide how many components to retain. We found that the first few dozen principal components accounted for the vast majority of variance in the 3{,}072-dimension embeddings. In fact, an analysis of explained variance showed a sharp elbow: by the time we include about 100–150 components, additional components contributed diminishing returns. Ultimately, we set the number of components to 110, which retained more than 50\% of the variance. This is a trade-off between loss of information and efficiency.

\begin{figure}[ht]
    \centering
    \includegraphics[width=0.7\textwidth]{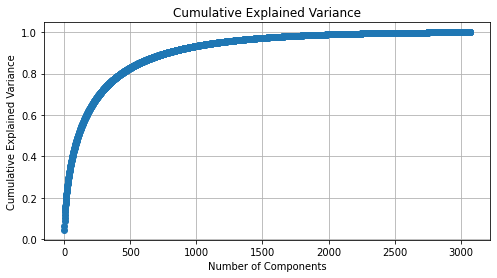}
    \caption{The cumulative explained variance of the PCA components.}
    \label{fig:pca-explained-variance}
\end{figure}

\begin{table}[ht]
    \centering
    \begin{tabular}{l c}
        \hline
        \textbf{Variance Threshold} & \textbf{Number of Components} \\
        \hline
        At least 50\%  & 105  \\
        At least 60\%  & 169  \\
        At least 70\%  & 266  \\
        At least 80\%  & 430  \\
        At least 90\%  & 774  \\
        At least 95\%  & 1159 \\
        At least 99\%  & 2025 \\
        \hline
    \end{tabular}
    \caption{Number of PCA components required to preserve specified levels of variance.}
    \label{tab:pca-variance}
\end{table}

After determining $n=110$, we transformed all sentence embeddings (train and test) into this PCA space. Each 3{,}072-dimension vector was thus converted into a 110-dimension vector. Importantly, this transformation is unsupervised, which means we are not tuning to our specific retrieval task labels but rather finding a general low-dimensional representation of the news data. The benefit is that the PCA projection could be learned on any large collection of unlabeled text embeddings (for instance, all news articles), making it widely applicable. It also means we are not overfitting to our particular ground truth; we are relying on the assumption that preserving variance in embeddings correlates with preserving semantic distinguishability for retrieval.

\begin{figure}[ht]
    \centering
    \includegraphics[width=0.7\textwidth]{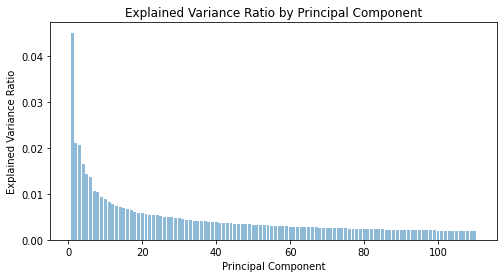}
    \includegraphics[width=0.7\textwidth]{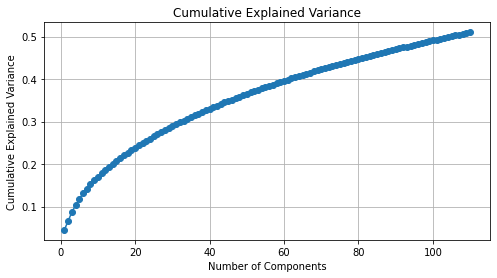}
    \caption{The cumulative explained variance of the PCA components till $n=110$.}
    \label{fig:pca-explained-variance}
\end{figure}

We then incorporated these reduced-dimensional embeddings into a retrieval pipeline. For the baseline approach, we use the original high-dimensional embeddings: given a query sentence embedding, we compute cosine similarity (and optionally L1 or L2 distance) against candidate sentence embeddings and rank the results. For the PCA-enhanced approach, we do the same except using the 110-dimension PCA vectors for both query and candidates when computing metrics.

To ensure fairness, all other aspects of the retrieval pipeline were identical between baseline and PCA-compressed versions. We index the same set of sentences, use the same similarity function, and evaluate on the same queries. The only difference is the vector representation. This allows us to isolate the effect of dimensionality reduction. Notably, by reducing vector size from 3{,}072 to 110 (a $\sim$28$\times$ reduction), the memory footprint of the index and the computation per lookup are drastically reduced. Storing an $N \times 3072$ embedding matrix vs.\ $N \times 110$ (for thousands or millions of sentences) is a significant difference. In a production setting like Newswitch processing streaming news, this translates to lower latency and cost for retrieval. Prior work emphasizes this efficiency gain: smaller dense vectors mean faster distance computations and less storage, at hopefully only a minimal trade-off in accuracy . Liu et al.\ (2022) specifically address dense retrieval efficiency, noting that high-dimensional embeddings “lead to larger index storage and higher retrieval latency,” and demonstrate that compression can maintain ranking performance while making retrieval more efficient. Our implementation leverages this insight by using PCA to achieve a more efficient RAG retrieval pipeline for news.

\subsection{Evaluation Framework}
We evaluated retrieval performance using both quantitative metrics and qualitative analysis. The evaluation is designed to answer: Does PCA compression hurt or help the ability to retrieve relevant information compared to using full embeddings?

\paragraph{Ground Truth and Task:} We treat each sentence in a pair as a query and the other sentence as the relevant document. Given the nature of our dataset (pairwise similarity scores), this calculates the dissimilarity between the two sentences and compares it against the scores given in the data frame.

\paragraph{Correlation with Similarity Scores:} Additionally, since our data has graded similarity scores (not just a binary relevant/irrelevant judgment), we evaluated how well the embedding similarities correlate with the human scores. We computed the Pearson correlation between the similarity functions produced by the model and the ground-truth similarity score \cite{benesty2009}, as well as the Spearman rank correlation between the model’s similarity ranking and the human ranking \cite{spearman1904}. A high Pearson correlation indicates that the model’s continuous similarity predictions align well with actual semantic similarity, and a high Spearman correlation indicates the model is good at ranking pairs in the correct order of relevance. For these calculations, each pair’s sentences were embedded, the cosine similarity was computed (using either full or PCA embeddings), and then correlated with the given score. We also calculated correlations for other distance metrics (L1, L2) for completeness.

\subsection{Distance and Similarity Metrics}

We employ four metrics to quantify the similarity (or distance) between sentence embeddings for measuring the accuracy and also the computation time for each of these:

\begin{itemize}
    \item \textbf{L1 Norm (Manhattan Distance)}:
    \[
    \mathrm{L1Norm}(u, v) \;=\; \sum_{i=1}^{n} \bigl| u_{i} \;-\; v_{i} \bigr|,
    \]
    where \(u, v \in \mathbb{R}^{n}\). This measures the “taxicab” distance between embeddings.

    \item \textbf{L2 Norm (Euclidean Distance)}:
    \[
    \mathrm{L2Norm}(u, v) \;=\; \sqrt{\sum_{i=1}^{n} \bigl(u_{i} \;-\; v_{i}\bigr)^{2}}.
    \]
    This is the standard Euclidean distance in \(\mathbb{R}^{n}\).

    \item \textbf{L1 Similarity}:
    \[
    \mathrm{L1Similarity}(u, v) \;=\;
    \frac{u \cdot v}{\sum_{i=1}^{n} \lvert u_{i}\rvert \;\sum_{i=1}^{n} \lvert v_{i}\rvert},
    \]
    where \(u \cdot v = \sum_{i=1}^{n} u_{i}v_{i}\). This metric normalizes the dot product by the L1 norm (sum of absolute values) of each vector.

    \item \textbf{Cosine Similarity}:
    \[
    \mathrm{CosineSimilarity}(u, v) \;=\;
    \frac{u \cdot v}{\|u\|\|v\|},
    \]
    where \(\|u\| = \sqrt{\sum_{i=1}^{n} u_{i}^{2}}\). Cosine similarity measures the cosine of the angle between the two embedding vectors.
\end{itemize}

All evaluations were done independently for the baseline and PCA-compressed systems. We then compared the metrics side by side. In summary, our framework covers both the effectiveness of retrieval and the consistency of similarity scoring with human judgment, providing a comprehensive view of performance.

\section{Results}

\subsection{Accuracy-Speed Trade-Off of PCA Compression}
To assess the impact of PCA-based dimensionality reduction, we computed the mean absolute error (MAE) between the ground-truth similarity scores and the distances/similarities produced by both full and PCA-compressed vectors. Table~2 and Table~3 summarizes the key findings. For Cosine Similarity, we observed a \(\sim 0.103\)-point increase in MAE when using PCA embeddings compared to the full 3{,}072-dimensional vectors, but achieved a roughly 60$\times$ speedup in similarity computation. Notably, L1 Similarity exhibited a smaller accuracy cost (\(\sim 0.041\)-point increase in MAE) while retaining the same 60$\times$ speedup. The two norm-based distances (L1~Norm, L2~Norm) yielded only a small (\(\sim 0.02\)) rise in MAE but showed lower gains in raw speed (about 2$\times$). 

\begin{table}[ht]
\centering
\begin{tabular}{l c}
\hline
\textbf{Metric} & \textbf{Time Performance (in Milliseconds)} \\
\hline
Cosine Similarity        & 0.5792 \\
L1 Similarity    & 0.5933 \\
L1 Norm        & 0.0148 \\
L2 Norm        & 0.0158 \\
Cosine Similarity - PCA   & 0.0099 \\
L1 Similarity - PCA & 0.0127 \\
L1 Norm - PCA   & 0.0056 \\
L2 Norm - PCA   & 0.0078 \\
\hline
\end{tabular}
\caption{Time performance for different distance/similarity metrics in Milliseconds.}
\label{tab:time-performance}
\end{table}

\begin{table}[ht]
\centering
\begin{tabular}{l c}
\hline
\textbf{Metric} & \textbf{Accuracy Performance (MAE)} \\
\hline
Cosine Similarity        & 0.1492 \\
L1 Similarity    & 0.1454 \\
L1 Norm        & 0.4257 \\
L2 Norm        & 0.4270 \\
Cosine Similarity - PCA   & 0.2524 \\
L1 Similarity - PCA & 0.1868 \\
L1 Norm - PCA   & 0.4416 \\
L2 Norm - PCA   & 0.4452 \\
\hline
\end{tabular}
\caption{Accuracy performance (MAE) for each metric.}
\label{tab:accuracy-performance}
\end{table}

These trends confirm that PCA compression introduces a clear trade-off between efficiency and retrieval fidelity. In practice, this trade-off is tunable by adjusting the number of retained principal components. As our PCA analysis suggested, retaining more components can recover additional variance and improve accuracy, albeit at the cost of slower inference and a larger index. Conversely, reducing to fewer components accelerates distance computations, which is valuable in real-time environments (e.g.\ streaming financial news and mid/high-frequency trading).

\subsection{Memory Efficiency and Index Size Reduction}
Another key benefit of PCA-driven compression lies in memory usage. Despite the modest size of our test corpus (8{,}600 rows), the PCA vectors consumed 28.6$\times$ less storage than the full 3{,}072-dimensional embeddings. Such a reduction is nontrivial when scaling to millions of vectors in production systems, such as retrieving news articles and regulatory filings. By storing fewer dimensions in the index, we reduce hardware and operational costs, making large-scale retrieval feasible under real-world constraints. These findings align with prior work \cite{ma2021,yang2021} demonstrating that compressed representations substantially lower index footprint without sacrificing too much retrieval accuracy.

\subsection{Distributional Analyses: Errors vs.\ Raw Metrics}

We conducted two complementary distributional analyses to better understand how PCA compression affects our retrieval metrics. First, we examined the \emph{absolute errors} of each distance/similarity metric relative to the annotated ground-truth scores. Second, we evaluated how PCA alters the \emph{raw} distributions of these metrics—i.e., the observed ranges and shapes of Cosine Similarity, L1 Similarity, L1 Norm, and L2 Norm when comparing sentence pairs.

\paragraph{Absolute Error Distributions:} Plotting the absolute errors in Figure~3 and comparing the distributions reveals that PCA typically shifts the error distributions slightly outward, reflecting the information loss from dimensionality reduction. For instance, both Cosine Similarity and L1 Similarity show peaks near zero absolute error for the full-dimensional embeddings, while the PCA versions exhibit slightly wider error curves. However, the majority of values remain clustered toward lower errors, indicating that most sentence pairs retain strong alignment with ground truth. Minimal “heavy tails” in the PCA distributions suggest that a small subset of pairs suffers increased distortion, but overall accuracy remains largely intact. Norm-based errors (L1 and L2) are similarly impacted, though their PCA distributions exhibit somewhat narrower ranges, suggesting that in some cases the PCA projection smooths out extreme distance values.

\begin{figure}[ht]
    \centering
    \begin{tabular}{cc}
        \includegraphics[width=0.45\textwidth]{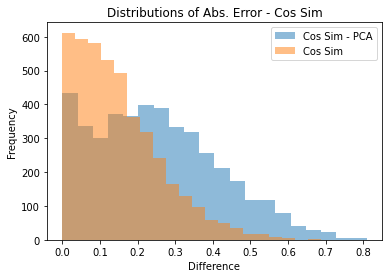} &
        \includegraphics[width=0.45\textwidth]{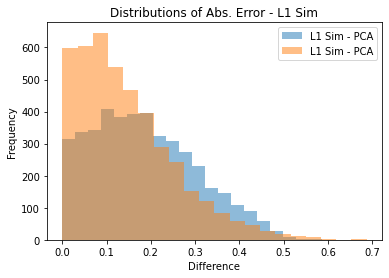} \\
        \includegraphics[width=0.45\textwidth]{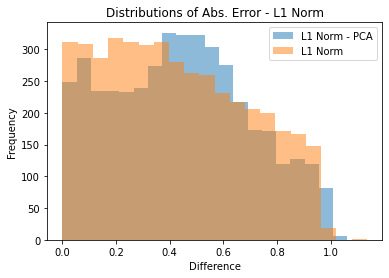} &
        \includegraphics[width=0.45\textwidth]{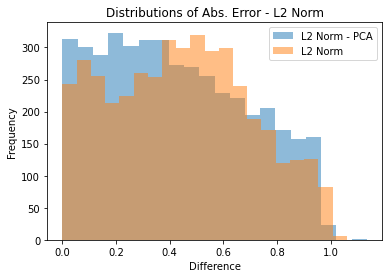} \\
    \end{tabular}
    \caption{Distribution of Absolute Errors for Distance and Dissimilarity Metrics.}
    \label{fig:pca-explained-variance}
\end{figure}

\paragraph{Raw Metric Distributions:} Turning to the raw metric values themselves, we observe, in Figure~4, a consistent narrowing of ranges under PCA for Cosine Similarity and L1 Similarity. Specifically, the PCA-compressed embeddings tend to compress the highest and lowest similarity scores closer to the mean, resulting in increased kurtosis (i.e., more pronounced central mass and lighter tails). This phenomenon aligns with PCA’s objective of preserving the dominant variance directions; noise or less informative directions get collapsed, leading to fewer extreme similarity values. Meanwhile, L1 Norm and L2 Norm distributions also become more concentrated under PCA, implying that distance outliers shrink toward the bulk of the distribution. Empirically, this can be beneficial in retrieval contexts, as extreme vector norms may reflect noise or irrelevant variability in the original embedding space.

\begin{figure}[ht]
    \centering
    \begin{tabular}{cccc}
        \includegraphics[width=0.22\textwidth]{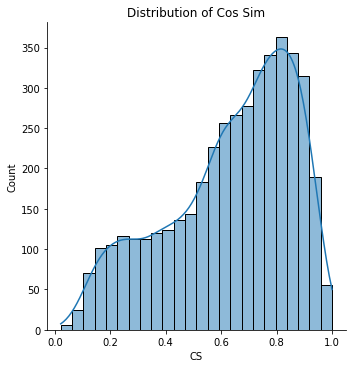} &
        \includegraphics[width=0.22\textwidth]{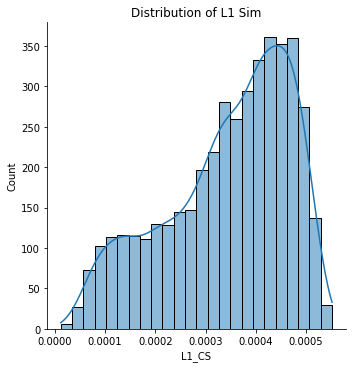} &
        \includegraphics[width=0.22\textwidth]{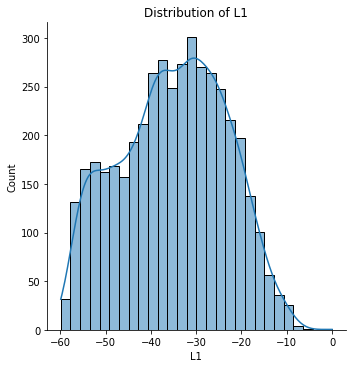} &
        \includegraphics[width=0.22\textwidth]{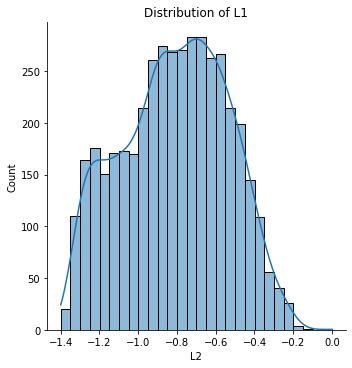} \\
        \includegraphics[width=0.22\textwidth]{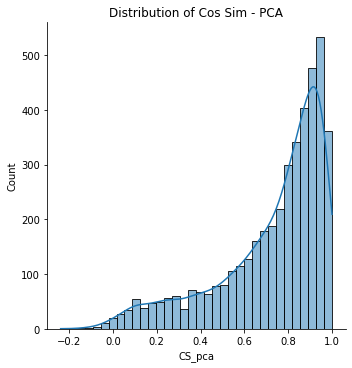} &
        \includegraphics[width=0.22\textwidth]{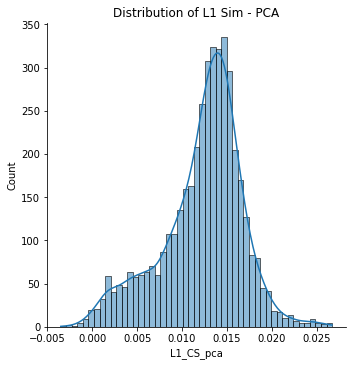} &
        \includegraphics[width=0.22\textwidth]{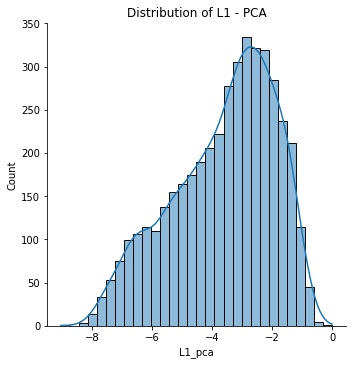} &
        \includegraphics[width=0.22\textwidth]{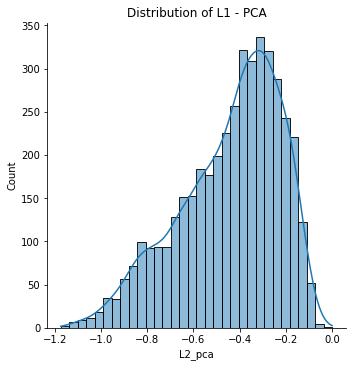} \\
    \end{tabular}
    \caption{Distribution of Raw Distance and Dissimilarity Metrics.}
    \label{fig:my-2x4-layout}
\end{figure}

\subsection{Ranking Consistency and Correlations}
To evaluate retrieval ranking consistency, we ranked unscaled metrics (Cosine, L1, and L2) against human-annotated scores. Because the gold-standard similarity ratings span only 90 unique values in \([0,1]\), we expect some “chunking” in the data. Nevertheless, QQ plots in Figure 5 revealed no drastic shifts for PCA-compressed embeddings: the ranked order of similarities largely mirrored that of the full-dimensional baseline. 

\begin{figure}[ht]
    \centering
    \begin{tabular}{cccc}
        \includegraphics[width=0.22\textwidth]{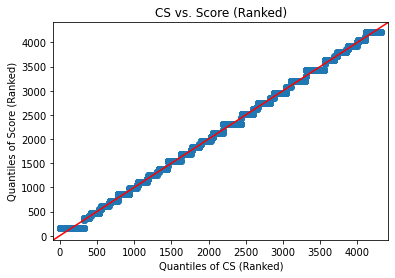} &
        \includegraphics[width=0.22\textwidth]{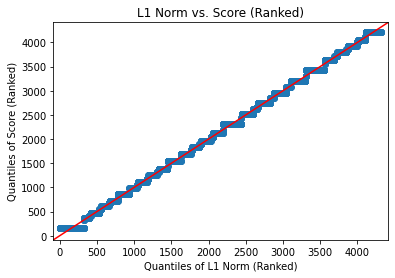} &
        \includegraphics[width=0.22\textwidth]{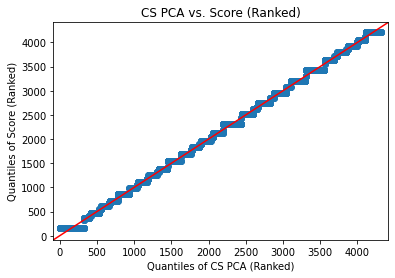} &
        \includegraphics[width=0.22\textwidth]{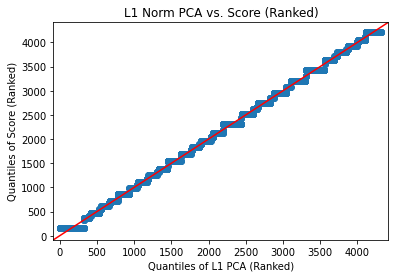} \\
    \end{tabular}
    \caption{QQ Plots of Ranked Cosine Similarity,  L1 Norm, and Their PCA vs. Score.}
    \label{fig:my-2x4-layout}
\end{figure}

We further examined the Pearson and Spearman correlations between the computed distances/similarities and the reference scores. Across all metrics (Cosine, L1, L2, and their variations), correlation coefficients remained comparable for PCA vs.\ full embeddings, indicating minimal to no loss in how well the models capture semantic relatedness. These results support the notion that preserving most of the variance with PCA suffices for robust correlation with human judgments.

\begin{figure}[ht]
    \centering
    \begin{tabular}{cccc}
        \includegraphics[width=0.45\textwidth]{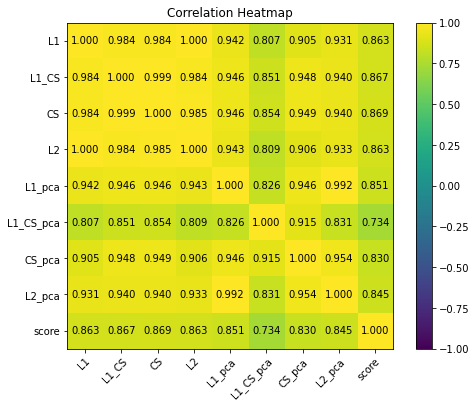} &
        \includegraphics[width=0.45\textwidth]{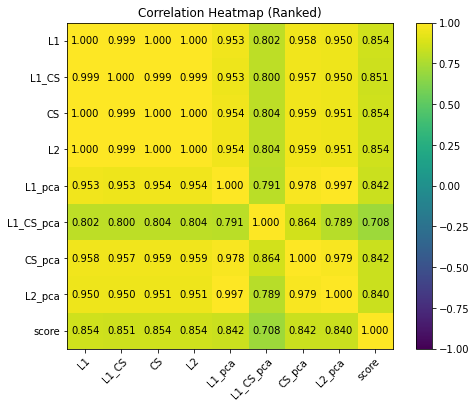} \\
    \end{tabular}
    \caption{Correlation Heatmap of Ranked and Raw Distance, Similarity Metrics, and Score.}
    \label{fig:my-2x4-layout}
\end{figure}

\subsection{Regression Analysis with Ground-Truth Scores}
To more formally quantify how each distance or similarity measure predicts the human-labeled scores, we performed a linear regression of the form 
\[
\text{Score} = \beta_0 + \beta_1 \times \text{(Distance/Similarity)} + \epsilon,
\]
where \(\beta_1\) is the slope. Table~4 reports the intercepts, slopes, \(t\)-statistics, \(p\)-values, coefficients of determination (\(R^2\)), and overall model significance. 

\begin{table}[ht]
\centering
\begin{tabular}{lrrrrrrrrrr}
\hline
\textbf{Predictor} & \textbf{Intercept}   & \textbf{Slope} & \textbf{t-stat} & \textbf{p-value} & \(\mathbf{R^2}\) & \textbf{F-stat}  \\
\hline
Cosine Similarity          & -0.2066  & 1.1489    & 115.39 & 0.0 & 0.7554 & 13315.24  \\
L1 Similarity      & -0.2022  & 2104.3703 & 114.21 & 0.0 & 0.7515 & 13043.49  \\
L1 Norm          & 1.3012    & 0.0222    & 112.02 & 0.0 & 0.7442 & 12548.12  \\
L2 Norm     & 1.2987    & 0.9575    & 112.42 & 0.0 & 0.7456 & 12637.60  \\
Cosine Similarity - PCA     & -0.2179    & 1.0162    & 97.66  & 0.0 & 0.6886 & 9537.19   \\
L1 Similarity - PCA & -0.0769   & 48.7458   & 71.02  & 0.0 & 0.5391 & 5044.53   \\
L1 Norm - PCA     & 1.0647    & 0.1507    & 106.49 & 0.0 & 0.7245 & 11339.80  \\
L2 Norm - PCA     & 1.0362   & 1.1668    & 103.58 & 0.0 & 0.7133 & 10728.38  \\
\hline
\end{tabular}
\caption{Regression coefficients, t-statistics, p-values, and performance measures.}
\label{tab:regression-results}
\end{table}

\begin{itemize}
    \item \textbf{Full Embeddings.} Cosine Similarity yielded the highest \(R^2\) (\(\sim0.755\)), closely followed by L1 Similarity(0.7515), L2 Norm (0.7456), and L1 Norm (0.7442). These metrics explain over 70\% of the variance in human scores, underscoring their suitability for semantic retrieval tasks. 
    \item \textbf{PCA-Compressed Embeddings.} Although R-squared values dipped slightly, PCA versions generally remained robust. For instance, Cosine Similarity - PCA attained \(R^2=0.6886\), while L2 Norm - PCA reached 0.7133, still capturing the bulk of variance in human judgments. The largest drop was for L1 Similarity - PCA (0.5391), suggesting that combining L1 normalization with Cosine in a compressed space is more susceptible to variance loss. 
    \item \textbf{Interpretation of Regression Parameters.} For full-embedding metrics, intercepts near zero (in Cosine-based approaches) or near one (in L1/L2 distances) reflect the baseline offset needed to align with average human scores. Slope values reveal how sensitively each metric scales with respect to the annotated similarity. While PCA modifies both intercept and slope, its explained variance remains within practical bounds, aligning with our correlation findings that the compressed vectors still track human similarity ratings closely.
\end{itemize}

Overall, these regression results reinforce the primary takeaway that PCA-based compression, at 110 principal components, preserves a substantial fraction of the explanatory power for semantic similarity. Although certain metrics (notably L1 Similarity) are more sensitive to dimensionality reduction, the total performance impact remains modest, especially considering the significant gains in speed and memory.

\subsection{Summary of Findings}
Our experiments show that PCA compression of 3,072-dimensional sentence embeddings to 110 dimensions yields:
\begin{enumerate}
    \item[i)] \textbf{Significantly Faster Inference.} 
    For Cosine and L1 Similarity, we observed up to a \(60\times\) reduction in distance computation time, 
    which is crucial for real-time or near-real-time retrieval scenarios.

    \item[ii)] \textbf{Reduced Memory Footprint.} 
    The PCA index consumes \(\sim 28.6 \times\) less storage, enabling more scalable deployment 
    when indexing millions of vectors.

    \item[iii)] \textbf{Minimal Accuracy Degradation.} 
    While error metrics (MAE, correlation, regression \(R^2\)) slightly worsen under PCA, 
    the deficits remain within acceptable bounds for many downstream applications—particularly 
    those willing to trade a small drop in fidelity for large gains in efficiency.

    \item[iv)] \textbf{Flexible Trade-Off.} 
    Adjusting the number of principal components offers a tunable balance between retrieval accuracy 
    and computational cost, supporting a range of real-world usage scenarios.
\end{enumerate}

In sum, these results confirm that PCA-driven dimensionality reduction is a viable strategy for speeding up dense retrieval in RAG pipelines. Despite modest losses in retrieval precision and correlation with human judgments, the memory and runtime savings are substantial, especially when scaling to large text corpora. Such efficiency benefits align well with practical requirements in finance-focused systems like Newswitch, where sub-second query response and high-volume index storage are critical.


\section{Limitations and Future Work}

\begin{enumerate}

    \item \textbf{Unsupervised Nature of PCA and Domain-Specific Concerns.}\\
    PCA selects directions of maximal variance without explicitly accounting for task- or domain-specific features \cite{abdi2010}. Although this captures much semantic information, critical nuances—especially in specialized domains like finance—may not align with the principal components. Outlier or niche financial terms could be overly compressed, impacting specialized retrieval tasks. As a result, future work should consider training PCA on domain-homogeneous corpora or employing domain-adaptive dimensionality reduction techniques \cite{gururangan2020,lee2020} to ensure that salient features are preserved.

    \item \textbf{Rank-Focused Metrics vs.\ Absolute Similarity.}\\
    While our experiments noted increases in absolute error metrics (e.g., MAE) after compression, the real impact in retrieval-based applications often depends on ranking quality rather than the exact distance values. If the correct documents remain within the top-$k$ neighbors, retrieval performance may be minimally affected. Incorporating rank-based metrics such as precision@$k$, recall@$k$, and mean reciprocal rank (MRR) \cite{voorhees1999} can thus offer a clearer picture of real-world system effectiveness.

    \item \textbf{Hybrid Compression Methods.}\\
    Although PCA provides a straightforward and interpretable baseline, more advanced approaches—such as product quantization (PQ) \cite{jegou2011}, transformer-based bottleneck layers \cite{tay2020}, or learned autoencoders \cite{liu2022}—may preserve finer-grained or domain-specific information. Combining PCA with these methods (e.g., PCA + PQ) could balance interpretability, compression ratio, and retrieval accuracy. Identifying optimal configurations under strict latency or memory constraints remains a promising area for further exploration.

    \item \textbf{Small, General-Purpose Dataset vs.\ Domain-Specific Training.}\\
    Our experiments used a moderately sized, general semantic similarity dataset. In realistic, large-volume financial or regulatory text collections, domain-adaptive PCA training may deliver better results by focusing on relevant domain features \cite{beltagy2019}. Future studies should assess whether training PCA on specialized corpora allows more aggressive compression without sacrificing retrieval fidelity.

    \item \textbf{Fixed Dimensionality and Adaptive Strategies.}\\
    We chose 110 principal components to strike a general balance between efficiency and accuracy. However, this fixed choice may be suboptimal for certain subdomains or specific query types. A more dynamic or query-adaptive compression mechanism \cite{li2022} could tailor the number of retained components to each context, improving resource utilization without compromising on crucial semantic detail.

    \item \textbf{Integration with Approximate Nearest Neighbor (ANN) Search.}\\
    ANN methods such as Hierarchical Navigable Small World (HNSW) graphs or Inverted File with Product Quantization (IVFPQ) \cite{johnson2019} can further accelerate retrieval, and their synergy with PCA-compressed embeddings warrants investigation. Optimizing top-$k$ performance and query speed in tandem—especially under real-time conditions—would help identify ideal end-to-end configurations for production systems like \emph{Newswitch}.

    \item \textbf{Deployment at Scale and End-to-End Validation.}\\
    Finally, large-scale deployment of a PCA-compressed RAG pipeline in high-throughput settings (e.g., live financial news retrieval) remains an important next step. Evaluating end-to-end performance—such as final answers, latency, and decision quality—under concurrency, evolving indices, and domain drift will clarify whether small distortions in embedding space translate into measurable degradation of outputs \cite{liu2023}.

\end{enumerate}

\noindent Overall, while PCA-based compression offers tangible benefits in speed and index efficiency, these considerations highlight the need to tailor dimensionality reduction to each domain’s requirements, preserve subtle yet crucial distinctions, and evaluate retrieval outcomes using metrics that align with real-world application goals.

\section{Conclusion}
Our investigation demonstrates that PCA-driven dimensionality reduction can substantially improve the efficiency of RAG pipelines without incurring large drops in retrieval effectiveness. Across varied similarity and distance metrics, reducing high-dimensional sentence embeddings to around 110 principal components yields a marked decrease in both computation time and index footprint \cite{ma2021, yang2021}. These gains are especially pertinent for finance and trading applications, where rapid responses to high volumes of news and market data are crucial \cite{iaroshev2024, xiao2025}.

Though we observed a modest increase in error metrics compared to full-dimensional embeddings, the overall drop in accuracy remained within an acceptable range for real-world usage, particularly given the scalability benefits. By reducing storage requirements and accelerating distance computations, PCA enables the construction of more responsive, cost-effective retrieval systems—such as Zanista AI’s \textit{Newswitch} platform—in which sub-second search results are essential to inform fast-paced trading decisions \cite{karpukhin2020, lewis2020}.

Future work may explore hybrid compression strategies, including learned autoencoders or product quantization, that build upon PCA’s efficiency gains \cite{liu2022, jegou2011}. Ultimately, our findings reaffirm that dimensionality reduction constitutes a vital component in the evolving landscape of retrieval-augmented architectures \cite{zhu2024, tay2020}, fostering more agile and robust solutions for knowledge-intensive domains like finance.

\end{document}